\documentclass[letterpaper, 10 pt, conference]{ieeeconf}  % Comment this line out if you need a4paper

\IEEEoverridecommandlockouts                              % This command is only needed if 
\usepackage{graphics} % for pdf, bitmapped graphics files
\usepackage{epsfig} % for postscript graphics files
\usepackage{mathptmx} % assumes new font selection scheme installed
\usepackage{times} % assumes new font selection scheme installed
\usepackage{amsmath} % assumes amsmath package installed
\usepackage{amssymb}  % assumes amsmath package installed
\usepackage{multirow}
\usepackage{cite} % for compressing consecutive citations

\usepackage{caption}
\DeclareCaptionFormat{hfillstart}{\hfill#1#2#3\par}
\DeclareCaptionFont{mdit}{\mdseries\itshape}
\usepackage{xcolor, soul, url}
\usepackage[font=small,skip=0pt]{caption}

\title{\LARGE \bf
Graph-SIM: A Graph-based Spatiotemporal Interaction Modelling for Pedestrian Action Prediction}
\vspace{-1cm}
\author{Tiffany Yau*$^\dagger$, Saber Malekmohammadi*, Amir Rasouli$^\ddag$, Peter Lakner, Mohsen Rohani, Jun Luo\\% <-this % stops a space
\vspace{-0.5cm}
\scalebox{.63}{
\texttt{tiffanyyk.yau@mail.utoronto.ca, \{saber.malekmohammadi, amir.rasouli, peter.lakner, mohsen.rohani, jun.luo1\}@huawei.com}
}
\thanks{The authors are with Noah's Ark Lab, Huawei, Canada.}
\thanks{*Equal contrib. $^\dagger$Work done as an intern at Huawei $^\ddag$Corresp. author}
}

\begin{document}

\setlength{\abovedisplayskip}{1pt}
\setlength{\belowdisplayskip}{1pt}

\maketitle
\thispagestyle{empty}
\pagestyle{empty}

%%%%%%%%%%%%%%%%%%%%%%%%%%%%%%%%%%%%%%%%%%%%%%%%%%%%%%%%%%%%%%%%%%%%%%%%%%%%%%%%
\begin{abstract}
One of the most crucial yet challenging tasks for autonomous vehicles in urban environments is predicting the future behaviour of nearby pedestrians, especially at points of crossing. Predicting behaviour depends on many social and environmental factors, particularly interactions between road users. Capturing such interactions requires a global view of the scene and dynamics of the road users in three-dimensional space. This information, however, is missing from the current pedestrian behaviour benchmark datasets. Motivated by these challenges, we propose 1) a novel graph-based model for predicting pedestrian crossing action. Our method models pedestrians' interactions with nearby road users through clustering and relative importance weighting of interactions using features obtained from the bird's-eye-view. 2) We introduce a new dataset that provides 3D bounding box and pedestrian behavioural annotations for the existing nuScenes dataset. 
On the new data, our approach achieves state-of-the-art performance by improving on various metrics by more than 15\% in comparison to existing methods.
The dataset is available at \url{https://github.com/huawei-noah/datasets/PePScenes}. %Upon publishing of this paper, our dataset will be made publicly available.
\end{abstract}
%%%%%%%%%%%%%%%%%%%%%%%%%%%%%%%%%%%%%%%%%%%%%%%%%%%%%%%%%%%%%%%%%%%%%%%%%%%%%%%%

\section{INTRODUCTION\vspace{-0.2\baselineskip}}

Predicting pedestrian behaviour is complex and requires a profound understanding of various environmental factors, particularly the interactions between traffic participants. 
In recent years, many behaviour prediction algorithms have been proposed \cite{Alahi_2016_CVPR, Bhattacharyya_2018_CVPR, Kooji_2014_ECCV, Zhao_2020_ArXiv, Yu_2020_ECCV, Haddad_2020_arXiv, Manh_2019_ISVC, Fang_2018_IV, Stip, Jaad}, most of which predict pedestrians' future trajectories \cite{Alahi_2016_CVPR, Bhattacharyya_2018_CVPR, Kooji_2014_ECCV, Zhao_2020_ArXiv, Yu_2020_ECCV, Haddad_2020_arXiv, Manh_2019_ISVC}. Some recently proposed algorithms model interactions of road users by encoding their dynamics information \cite{Yu_2020_ECCV, Haddad_2020_arXiv, Stip, Sun_2020_CVPR, Mohamed_2020_CVPR}. However, these methods mainly rely on road users' trajectories without capturing information like traffic direction and group behaviours, key factors to modelling interactions in traffic context. 

There are a number of datasets that cater to pedestrian crossing prediction \cite{Stip}, \cite{Jaad}, \cite{Pie}, \cite{Titan}, \cite{Viena2}.  They contain video clips annotated with 2D bounding boxes and behavioural tags for pedestrians. These datasets, however, lack some key features, such as bird's-eye-view maps of environments and 3D locations of objects, which are necessary for capturing the interactions between road users.

In this work, we propose a pedestrian crossing prediction model that consists of a novel graph-based technique for modelling the interactions between road users by leveraging their coordinates, orientations, and lane information from a bird's-eye-view perspective. Using the interaction model in conjunction with the ego-vehicle and target pedestrian dynamics information, our model predicts the probability of crossing action. 

Furthermore, we introduce a new behaviour prediction dataset generated by adding dense 3D bounding boxes and behavioural annotations to the existing nuScenes dataset \cite{Caesar_2020_CVPR}.  Using the new dataset, we train our proposed model and compare its performance with existing pedestrian behaviour prediction frameworks.

\begin{figure}[tp]
    \begin{center}
    \includegraphics[width = 1\columnwidth]{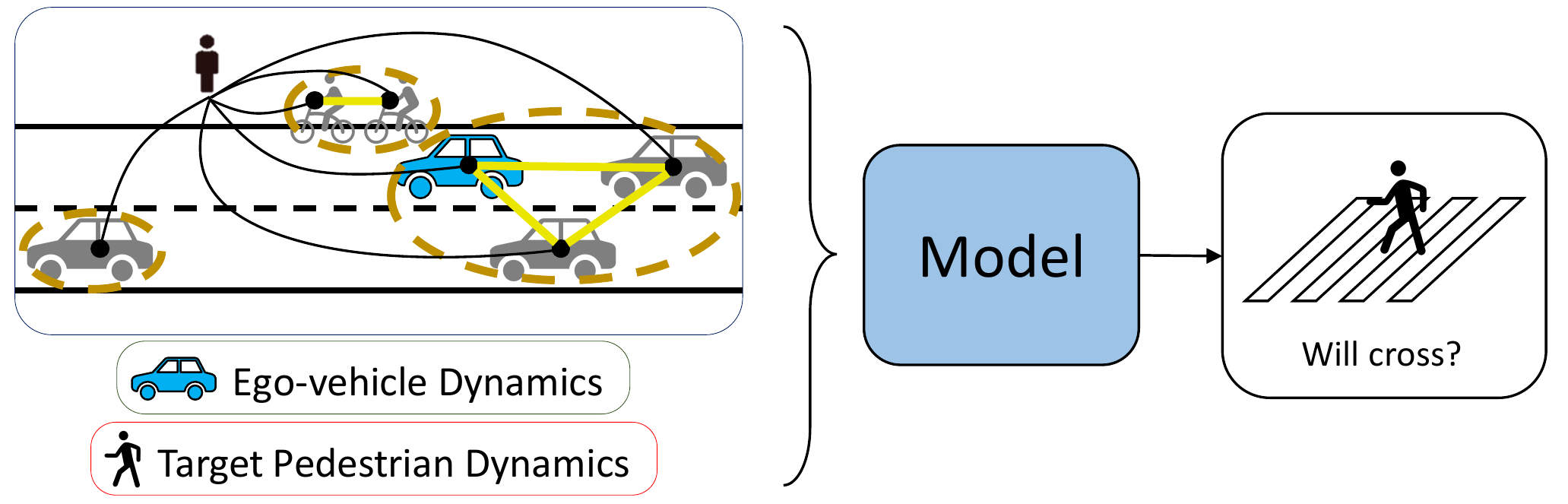}
    \vspace{-1em}
    \caption{The proposed pedestrian action prediction method models the spatiotemporal interactions of a pedestrian and its surrounding road users. This information, combined with ego-vehicle and the pedestrian's dynamics, is used to predict whether the pedestrian will cross the road.
    \label{fig:pap}
    \vspace{-1.5em}}
    \end{center}
\end{figure}

\section{RELATED WORKS \vspace{-0.2\baselineskip}}
\subsection{Pedestrian Behaviour Prediction\vspace{-0.2\baselineskip}}
The common approach to behaviour prediction is forecasting trajectories based on observed dynamics and other contextual information \cite{Alahi_2016_CVPR, Mohamed_2020_CVPR, Sun_2020_CVPR_2, Kosaraju_2019_NIPS, Zhang_2020_CVPR, Park_2020_ECCV}. 

Another approach to behaviour prediction is to anticipate future actions, such as crossing the road \cite{Jaad, Chaabane_2020_WACV} or interactions with other objects \cite{Liang_2019_CVPR}. In the context of driving, crossing prediction is particularly important for safe motion planning. In recent years, a number of methods have been proposed for pedestrian crossing prediction. A group of these methods rely on feedforward approaches, in which predictions are made based on changes in the scene images \cite{Gujjar_2019_ICRA, Saleh_2019_ICRA, Chaabane_2020_WACV,Jaad}. For example, the methods in \cite{Gujjar_2019_ICRA,  Chaabane_2020_WACV} employ a generative approach to predict future scene images, which are used to predict future crossing events. The authors of \cite{Saleh_2019_ICRA} use a 3D DenseNet architecture that detects and predicts pedestrian actions in a single framework.

Recurrent approaches, which rely on multiple data modalities, are also widely used \cite{Stip, Rasouli_2019_arXiv,Viena2}. For instance, the method of \cite{Rasouli_2019_arXiv} uses a multi-layer recurrent architecture that processes different data modalities including pedestrian appearance and their surroundings, poses, trajectories, and ego-vehicle speed. Each data modality is fused into the network according to their complexity. The authors in \cite{Stip} present a graph-based method, which uses 2D image representations of traffic objects as nodes and relative distance to assign weights to edges. The graph is processed using a GCNN followed by a GRU and a dense layer to classify future actions.

The existing pedestrian crossing models only rely on 2D image information, which is insufficient for autonomous driving applications. In our proposed approach, we take advantage of 3D information to reason about the dynamics of the objects in the bird's-eye-view perspective.

\subsection{Methods of Pedestrian Interaction Modelling \vspace{-0.2\baselineskip}}
Modelling pedestrians' interactions with their surroundings is one of the key components in behaviour prediction. The works in \cite{Alahi_2016_CVPR, Gupta_2018_CVE, Sun_2020_CVPR_2, Mangalam_2020_ECCV} use social pooling to associate equal importance to the objects in the neighbourhood around the pedestrian. In \cite{Vemula_2018_ICRA, Sadeghian_2019_CVPR, Park_2020_ECCV}, the methods use attention-based models, which associate relative importance to objects using their distances to pedestrians. 

Recent approaches use graph-based architectures to model interactions \cite{Haddad_2020_arXiv, Mohamed_2020_CVPR, Kosaraju_2019_NIPS, Zhang_2020_CVPR}. These methods build graphs, where nodes are represented by pedestrians and the relative importance is calculated based on the distance. A common disadvantage of the methods is that they assume the closer the objects to pedestrians, the more important they are. However, in reality, the direction of motion is also important. For example, if a vehicle is moving away from a pedestrian, it is not as important as an approaching vehicle, even within the same distance from the pedestrian. In the proposed approach, we calculate the relative importance by considering the distance as well as the location and traffic direction, and further encode the orientation of objects in node representations.

\subsection{Existing Datasets\vspace{-0.2\baselineskip}}

There are only a few datasets for crossing action prediction: Joint Attention in Autonomous Driving (JAAD) \cite{Jaad}, Pedestrian Intention Estimation (PIE) \cite{Pie}, Trajectory Inference using Targeted Action priors Network (TITAN) \cite{Titan}, VIrtual ENvironment for Action Analysis (VIENA\textsuperscript{2}) \cite{Viena2}, and Stanford-TRI Intent Prediction (STIP) \cite{Stip}. These datasets have rich behavioural tags for various pedestrian actions with temporally coherent spatial annotations. However, the datasets lack information such as, semantic maps of the environment, 3D coordinates of objects, etc. that are important for autonomous driving applications.  We address these shortcomings by introducing a novel dataset that is generated by adding dense 3D bounding box and behavioural annotations to the existing nuScenes dataset \cite{Caesar_2020_CVPR}.

\textbf{Contributions.} The contributions of this work are as follows: 1) We propose a novel graph-based interaction modelling approach which takes road users' bird's-eye-view information as input. Unlike the existing methods, the method assigns importance weights to different road users based on both their distance and their relative location with respect to a pedestrian. 2) We introduce a novel dataset, created by adding new per-frame bounding boxes and behavioural annotations to the nuScenes dataset \cite{Caesar_2020_CVPR}, at a higher frequency of 10 Hz. It leverages the benefits of the map, LIDAR, and camera data from nuScenes for pedestrian prediction. 3) We compare our method to state-of-the-art graph modelling \cite{Mohamed_2020_CVPR} and pedestrian crossing prediction \cite{Rasouli_2019_arXiv} methods and show how the proposed method achieves improved results. 4) Through our ablation studies, we highlight the contribution of each component of our method to the prediction task.

\section{Method Description\vspace{-0.2\baselineskip}}

% keeping track of notation
% A - Adjacency, 
% B - relative importance matrix,
% C - clusters, 
% D - distance matrix, (d - distance)
% \vec{e} - ego-vehicle dynamics, 
% F - frames, 
% G - graphs, 
% k - probability of crossing at frame t+k (problem formulation)
% L - locations, 
% N - #obj per traj., 
% O - orientations, 
% P - target pedestrian output encoding from graph, 
% \vec{p} - pedestrian dynamics,
% Q - # features, \Tilde{Q} - # output features from graph
% S - speeds, 
% T - sequence length (5), 
% V - node representations, 
% Y - event of crossing, used in probability distribution formulation
% Z - output of spatial convolution, Z = AVW equation
% i/j/m/n - arbitrary indices for equations
% remaining letters: H, R, U, W, X

\subsection{Problem Formulation\vspace{-0.2\baselineskip}}
We formulate pedestrian action prediction as an optimization problem. The goal is to estimate the probability distribution of crossing action %,
in the path of the ego-vehicle,
$p(Y_{t+k}^{(i)}|L_o, O_o, R_o) \in [0,1]$, where $Y_{t+k}^{(i)} \in \{0,1\}$, for the $i^{th}$ pedestrian at some time $t+k$ in the future, given an observed sequence of 
% speeds $S_o=\{s_1,s_2,...s_T\}$ and 
global bird's-eye-view locations $L_o=\{l_1,l_2,...,l_T\}$ and orientations $O_o=\{o_1,o_2,...,o_T\}$ of the target pedestrian, ego-vehicle, and nearby road users, along with traffic directions of the road $R_o=\{r_1,r_2,...,r_T\}$. 

\begin{figure*}[htp]
    \includegraphics[width=1\textwidth]{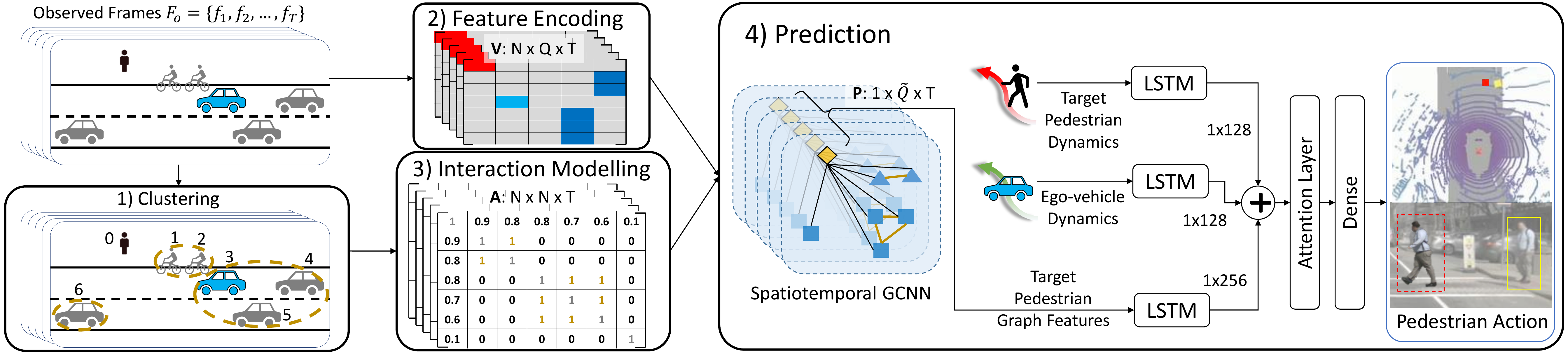}
    \vspace{-0.5em}
    \caption{The overview of our proposed model, Graph-SIM (Graph-based Spatiotemporal Interaction Modelling), which consists of four steps. 1) We cluster road users close to the target pedestrian in each frame. 2) Using bird's-eye-view information, we form feature encodings, which are used as graph node representations. 3) We assign weights to the graph edges (interactions) using the clusters and a relative importance calculation. 4) The graphs are fed to a spatiotemporal GCNN, the outputs of which are combined with the pedestrian and ego-vehicle dynamics and fed to three LSTMs, followed by an attention layer, and a dense layer for prediction.}
    \vspace{-1.8em}
    \label{fig:architecture}
\end{figure*}

\subsection{Architecture\vspace{-0.2\baselineskip}}
\label{architecture}
Our proposed spatiotemporal graph-based method models road users' interactions. It contains four key parts:
\begin{itemize}
    \item \textbf{Clustering.} Here, we group road users near the target pedestrian using their speed, location, and orientation.
    \item \textbf{Feature Encoding.} We form a vector of features for each object and the target pedestrian. The vectors are used as node representations of the graph.
    \item \textbf{Modelling Interactions via Graphs.} We calculate the relative importance of road users to the target pedestrian and use it with the clusters to weight the graph edges.
    \item \textbf{Prediction.} Using the graph representations and some dynamics information of the target pedestrian and ego-vehicle, our model estimates the probability of crossing action in the future.
\end{itemize}
We discuss each component in more detail in the following sections. A diagram of our method is shown in Figure \ref{fig:architecture}. 

\subsection{Clustering\vspace{-0.2\baselineskip}}
\label{clustering}
Objects interacting in groups in a traffic scene can exhibit similar behaviours \cite{Rasouli_2019_ITS}. As such, we model these interactions by doing a frame-wise clustering of the target pedestrian's surrounding road users using their static and dynamic properties, including object type, speed, location, and orientation. 

\textbf{Object types.} First, we separate the road users into three classes -- pedestrians, vehicles, and bicycles. We then further cluster objects within each class for every frame in a sequence of observations $F_o=\{f_1,f_2,...,f_T\}$, first based on speed $S_o=\{s_1,s_2,...s_T\}$, then bird's-eye-view locations $L_o$, and finally orientations $O_o$, in a hierarchical manner, obtaining a set of clusters for each observed frame, $C_o=\{c_1,c_2,...,c_T\}$.

\textbf{Object speed.} Within each of the three classes, we determine whether each object is moving or stationary by calculating its speed between the previous and current frame. For the $i^{th}$ object in frame $t$,
\begin{equation}
    \label{eq:speed}
    s_{t}^{(i)}=\frac{\left\|l_{t}^{(i)}-l_{t-1}^{(i)}\right\|}{frame \, rate}
\end{equation}
where $t=2...T$ and $frame\, rate$ is the frequency in Hz of the observations. At $t=1$, we set $s_{1}^{(i)}=0$. 

To split each object class into moving or stationary subgroups, we empirically set speed thresholds of 0.2 m/s, 2 m/s, and 2 m/s for pedestrians, vehicles, and bicycles respectively. If $s_{t}^{(i)} \geq threshold$, object $i$ is considered moving in frame $f_t$. Thus, all objects in a frame are divided into six groups: \textit{stationary} or \textit{moving} \textit{pedestrians}, \textit{vehicles}, or \textit{bicycles}.

\textbf{Locations.} Next, we generate coarse clusters within each group of objects based on their distances from each other in the bird's-eye-view. The Density-Based Spatial Clustering of Applications with Noise (DBSCAN) algorithm \cite{Ester_1996_KDD} is used for this. DBSCAN requires two parameters -- the maximum distance for two points to be considered in the same cluster, and MinPts, used to determine noise in clustering. We set MinPts to 1 so that no object is identified as noise. We empirically choose the maximum distance to be 1.5, 10, and 5 meters for pedestrians, vehicles, and bicycles respectively.

\textbf{Orientations.} We further split these distance clusters based on the objects' orientation information. For stationary pedestrians, no further clustering is done due to the highly variant nature of pedestrian orientations in stationary groups. 

With moving pedestrians, clustering is done based on two factors -- 1) whether two given pedestrians are facing in the same or different directions and 2) whether they are moving towards or away from each other. 

We consider two pedestrians, objects $i$ and $j$, to be facing opposite directions in frame $f_t$ if the angle between their orientation vectors, $\gamma$, satisfies 
% \small
\begin{equation}
    \gamma=\cos^{-1}{\left( \frac{\Vec{o_{t}^{(i)}} \cdot \Vec{o_{t}^{(j)}}} {\left\|\Vec{o_{t}^{(i)}}\right\| \left\|\Vec{o_{t}^{(j)}}\right\|} \right)} \geq 90^{\circ},
\end{equation}
% \normalsize
where $\Vec{o_{t}^{(i)}}$ and $\Vec{o_{t}^{(j)}}$ are the orientation vectors. Otherwise, we consider them to be facing the same direction. 

Two pedestrians are considered to be moving away from each other if the distance between them increases from frame $f_{t-1}$ to $f_t$. That is, 
\begin{equation}
    \left\|l_{t}^{(i)}-l_{t}^{(j)}\right\| > \left\|l_{t-1}^{(i)}-l_{t-1}^{(j)}\right\|.
\end{equation}
Otherwise, we consider them moving towards each other. 

We keep two pedestrians in a distance cluster together if they are: 1) facing the same direction, or 2) facing opposite directions and moving towards each other. We separate two pedestrians in the distance cluster if they are facing opposite directions and moving away from each other. 

For vehicle and bicycle clusters, we split them into two groups using the K-means algorithm \cite{Lloyd_1982_IT}, clustering based on their normalized orientation vectors. The goal of doing this is to prevent vehicles and bicycles travelling in different directions to be placed in the same cluster.

We use the set of clusters found in frame $f_t$, $c_t=\{c_{t}^{(1)},c_{t}^{(2)},...,c_{t}^{(N_{C,t})}\}$, where $N_{C,t}$ is the total number of clusters at frame $f_t$, for weighting graph edges (Section \ref{adjacency}).

\subsection{Feature Encoding\vspace{-0.2\baselineskip}}
\label{features}
Each road user in the traffic scene is represented as a node in the graph for a given frame. To form the node representation for the $i^{th}$ object in frame $f_t$, we form a vector, $\Vec{v}_{t}^{(i)}$, containing 35 values. The vector includes information on its object type, location, motion, and size. As opposed to the feature representation in \cite{Mohamed_2020_CVPR} which only uses bird's-eye-view coordinates, we include additional information to better represent each object.

For each object, seven out of the 35 elements of $\Vec{v}_{t}^{(i)}$ directly pertain to the object's features. %are informative values. 
Of these seven elements, the first two represent whether the object is stationary $[1,0]$ or moving $[0,1]$. Including this allows the spatial graph convolution operation to encode information on the ratio of stationary to moving objects' importance to the pedestrian. 

The third and fourth elements represent an object's location with respect to the target pedestrian. For the target pedestrian, these values will be $[0,0]$. Otherwise, for object $i$, these values are $[x,y]=l_{t}^{(i)}-l_{t}^{(ped)}$. We consider a threshold $d_{thresh}=20m$, set empirically, around the target pedestrian. Thus, we adjust these coordinates to range from 0 to 1 using
\begin{equation}
    \label{eq:d_thresh}
    d=\frac{min(d,d_{thresh})}{d_{thresh}} 
\end{equation}
where $d$ is the distance of an object to the target pedestrian along the global $x$ or $y$ axis.

The fifth element is the speed of the object, as defined in Equation \ref{eq:speed}. The sixth and seventh elements are the object's length and width respectively. We find the maximum values of the speed, length, and width, and rescale the values in the training set to range between 0 and 1. The same factors are used to scale the values in the test set to a similar range.

After obtaining the seven values for an object, specific sections of $\Vec{v}_{t}^{(i)}$ are populated based on the object's type -- target pedestrian, ego-vehicle, other pedestrians, other vehicles, and bicycles. Thus, we sparsify $\Vec{v}_{t}^{(i)}$ into five sections of seven elements each. 

If the node represents the target pedestrian, we populate the first seven values of $\Vec{v}_{t}^{(i)}$. For the ego-vehicle, we populate the next seven values. For other pedestrians, other vehicles, and bicycles, we populate the third, fourth, and fifth sections respectively. The purpose of separating the features of different object types is to help the model identify the types of objects surrounding the target pedestrian.

For $N$ unique objects in a sequence, we stack the feature vectors into a matrix of size $N \times Q$ for each frame, where $Q=35$ is the length of $\Vec{v}_{t}^{(i)}$. Thus, we obtain feature matrices $V_o=\{V_1,V_2,...,V_T\}$. In preparation for the spatiotemporal convolution, the feature matrices are stacked into a 3D tensor of size $N \times Q \times T$.

\subsection{Modelling Interactions via Graphs\vspace{-0.2\baselineskip}}
\label{adjacency}
We construct graph structures for each frame of a target pedestrian's observed sequence, using the feature vectors as node representations and the generated clusters to form a symmetric adjacency matrix. 

Inspired by the pedestrian-centric star graph representation presented in \cite{Stip}, we first connect the target pedestrian node with the node of each road user appearing in the scene within a frame, $f_t$. Then, based on the clusters formed at $f_t$, $c_t=\{c_{t}^{(1)},c_{t}^{(2)},...,c_{t}^{(N_{C,t})}\}$, we connect objects to every other object in its own cluster to form a fully-connected subgraph.

The adjacency matrix $A_t$, at frame $f_t$ represents these node connections or graph edges. It is determined by two matrices -- $B_t$, a matrix encoding the relative importance of road users to the pedestrian, and $D_t$, a matrix representing the distance between the pedestrian and each object. For consistency, we always represent the pedestrian at the $0^{th}$ row and column of the adjacency matrix.

\textbf{Relative importance matrix, B.} For objects on a drivable area of the scene, the elements of $B$ are generated using its distance to the target pedestrian, calculated along the road closest to the pedestrian, in the direction of traffic. This distance measurement is robust to the driving direction of lanes and the curvature of the road. 
We assign a negative value to the distance if the object has passed the pedestrian and a positive value if the object is approaching. 

We empirically set a distance threshold, $d_{thresh}$, of 20 meters for the maximum distance from the target pedestrian we consider, and normalize values accordingly. Thus, for the $i^{th}$ object in $f_t$, which is on a drivable area of the scene, at a distance $d$, along the lane, we set
\begin{equation}
    B_{t}[0,i]=B_{t}[i,0]=\frac{min(max(d,-d_{thresh}),d_{thresh})+d_{thresh}}{2 \times d_{thresh}}.
\end{equation}
% \normalsize
For any object, $j$, on a non-drivable area, we set $B_{t}[0,j]=B_{t}[j,0]=0.5$.

For any two objects, $m$ and $n$, that are part of the same cluster, we set $B_{t}[m,n]=B_{t}[n,m]=0$. We assign 0 to all diagonal elements of $B_{t}$, and the rest are set to 1.

\textbf{Distance matrix, D.} The elements of $D$ are calculated using the Euclidean distance between the target pedestrian and its surrounding road users. Thus, for the $k^{th}$ object in $f_t$,
\begin{equation}
    d=\left\|l^{t}_{(0)}-l^{t}_{(k)}\right\|.
\end{equation}

Similar to the $B$ matrix, we use the distance threshold, $d_{thresh}=20 m$, to normalize the distances. 
To obtain $D_{t}[0,k]$ and $D_{t}[k,0]$, Equation \ref{eq:d_thresh} is used. 

As with $B$, for any two objects, $m$ and $n$, that are part of the same cluster, we set $D_{t}[m,n]=D_{t}[n,m]=0$. We also assign 0 to all diagonal elements of $D_{t}$. For all other elements of $D_{t}$, we assign a value of 1.

\textbf{Adjacency matrix, A.} We use the equation 
\begin{equation}
    A_t=(1-B_{t})\odot (1-D_{t})
\end{equation}
(where $\odot$ denotes element-wise multiplication) to obtain the adjacency matrix at frame $f_t$.

In summary, the value of $A_t[i,j]$ for the $i^{th}$ and $j^{th}$ objects in the scene at frame $f_t$, will be 1 when $i=j$ or when the objects belong in the same cluster. For all edges between the target pedestrian and each object, $0 \leq A_t[i,j] \leq 1$. When the two objects are not part of the same cluster,  $A_t[i,j]=0$. 

In preparation for the spatiotemporal convolution of the graph, we format the adjacency matrix such that for an observed sequence with a total of $N$ unique surrounding objects across all frames, the adjacency matrix will have dimension $N \times N$ at each frame in the sequence. 

If an object in the sequence is not present in a given frame, we do not connect it to any other node. This is done by setting all values of the adjacency matrix to zero for the corresponding rows and columns, except for the diagonal element, which is set to 1. The adjacency matrices at all frames of a target pedestrian's observed sequence, $A_o=\{A_1,A_2,...A_T\}$, are stacked into a 3D tensor of dimension $N \times N \times T$ and used in the spatiotemporal graph convolution.

\subsection{Prediction\vspace{-0.2\baselineskip}}

\textbf{Spatiotemporal Graph Convolution Neural Network.}
A spatial graph convolution is defined as $Z=AVW$ \cite{kipf}, where A is the adjacency matrix at one frame, V is the corresponding feature matrix, and W contains the trainable weights of the spatial convolution layer. Extending upon this, as in \cite{Mohamed_2020_CVPR}, a spatiotemporal graph involves constructing a graph using the set of spatial graphs from the observed frames $G_o=\{G_1,G_2,...,G_T\}=(V_o,A_o)$. These graphs have the same configuration at each frame, while the edge weights in $A_t$ and features in $V_t$ vary as $t$ ranges from $1$ to $T$. 

In the spatiotemporal graph convolution component, we leverage the clustering information by using two layers of spatial convolution, thus incorporating a level of each object's indirect neighbours into the convolution. With the convolution in the temporal dimension, we use a kernel size of 3, combining a given frame's information with that of its previous and next frames. For this component of our model, we use the PReLU \cite{Prelu} activation function, as in \cite{Mohamed_2020_CVPR}.

We denote the target pedestrian's embeddings as $P$, which has dimensions $1 \times \Tilde{Q} \times T$. In our implementation, we empirically selected the hyperparameter $\Tilde{Q}=512$ as the graph's output dimension. 

\textbf{Encoding Target Pedestrian and Ego-Vehicle Dynamics.}
\label{sect:dynamics}
To directly capture information pertaining to target pedestrian and ego-vehicle dynamics, we also encode a vector of pedestrian features, $\vec{p_t}=[p_{x,t},p_{y,t},p_{vx,t},p_{vy,t}]$, and a vector of ego-vehicle features, $\vec{e_t}=[e_{x,t},e_{y,t},e_{vx,t},e_{vy,t}]$, at each frame $f_t$. Here, $x$ and $y$ are the locations of the pedestrian and ego-vehicle and $vx$ and $vy$ are the velocities in the global bird's-eye-view reference frame. Velocity is calculated as $[vx, vy]=(l_{t}^{(i)}-l_{t-1}^{(i)})$, for $t=2...T$ where $i$ is the target pedestrian or ego-vehicle. At $t=1$, we set $[vx,vy]=[0,0]$. We also multiply the velocity by 1000 to scale it to an order of magnitude that is similar to the global $x$ and $y$ coordinates. 

\textbf{Action Classification.} Finally, for prediction, we feed the target pedestrian's graph embeddings $P$, the pedestrian dynamics $\{\vec{p_1},\vec{p_2},...,\vec{p_T}\}$, and the ego-vehicle dynamics $\{\vec{e_1},\vec{e_2},...,\vec{e_T}\}$ to three LSTMs. The hidden states are concatenated and fed to an attention layer, which weights the importance of the recurrent features. This is followed by a dense layer for classification. For training, we use binary cross-entropy loss.

\section{Dataset\vspace{-0.2\baselineskip}}
We introduce our dataset, a set of bounding box and behavioural annotations that extends upon the nuScenes dataset \cite{Caesar_2020_CVPR}. We refer to this dataset as Pedestrian Prediction on nuScenes (\textbf{PePScenes}). In addition to pedestrian action prediction, the added bounding boxes can be used in numerous tasks including detection, tracking, trajectory prediction. 

The nuScenes dataset has 1000 segments in total, 850 of which have available 3D bounding box annotations with temporal correspondence as part of the train and validation set. We add bounding box annotations for all labelled objects in nuScenes, as discussed in Section \ref{spatial_ann}. Additionally, we add behavioural annotations to a subset of pedestrian samples, as discussed in Section \ref{action_label}.

\subsection{Spatial Annotations\vspace{-0.2\baselineskip}}
\label{spatial_ann}

\subsubsection{Bounding Box Interpolation}
nuScenes contains LIDAR sweeps and camera frames taken at 20 and 12 Hz respectively, with bounding box annotations provided at 2 Hz. Since this is quite sparse, particularly for pedestrian behaviour prediction, we augment the spatial annotations of all existing objects to 10 Hz. We interpolate the translation, rotation, and size of the original bounding boxes between two consecutive original annotations. 
One of the 20 Hz LIDAR sweeps is associated with every new annotation, selected to be evenly spaced between two consecutive 2 Hz sweeps. We also associate a frame from each camera captured at the closest point in time to the corresponding LIDAR sweep.

\subsubsection{Verification of Interpolation}
To verify the alignment of the new bounding boxes with the actual objects, especially those in front of the ego-vehicle, we used a 2D detection algorithm, RetinaNet \cite{Lin_2017_CVPR}, pre-trained on MSCOCO \cite{Lin_2014_ECCV}, to localize objects in the front image frames. 
For pedestrians, based on each detection IoU, we selectively visualized the interpolated boxes and adjusted them manually when required.
We randomly sub-sampled 20\% of the other objects and evaluated them to ensure the quality of the new annotations.

\subsection{Action Labels\vspace{-0.2\baselineskip}}
\label{action_label}
Behavioural labels for pedestrian crossing actions were added to a subset of the pedestrians, selected based on whether they 1) appear in front of the ego-vehicle, 2) are not far from the drivable area, and 3) if they cross, are observable for at least a few frames prior to crossing. Following these criteria, a total of 719 unique pedestrians are selected, 149 of which are crossing pedestrians. We give each of these pedestrians an object-level annotation indicating whether they will cross the road in front of the ego-vehicle. 

Additionally, for crossing pedestrians, we include labels to specify the critical frame at which crossing starts. If the pedestrian is visible in a front camera frame when it finishes crossing, we also label this. We use this information to provide frame-wise crossing state labels for each pedestrian when visible in a front camera. For all frames between starting and ending, the pedestrian is labelled to be crossing. 
In total, we provide 63.4k per-frame behavioural annotations. 

\begin{figure}[htp]
    \centering
    \vspace{-0.25em}
    \includegraphics[width=1\columnwidth]{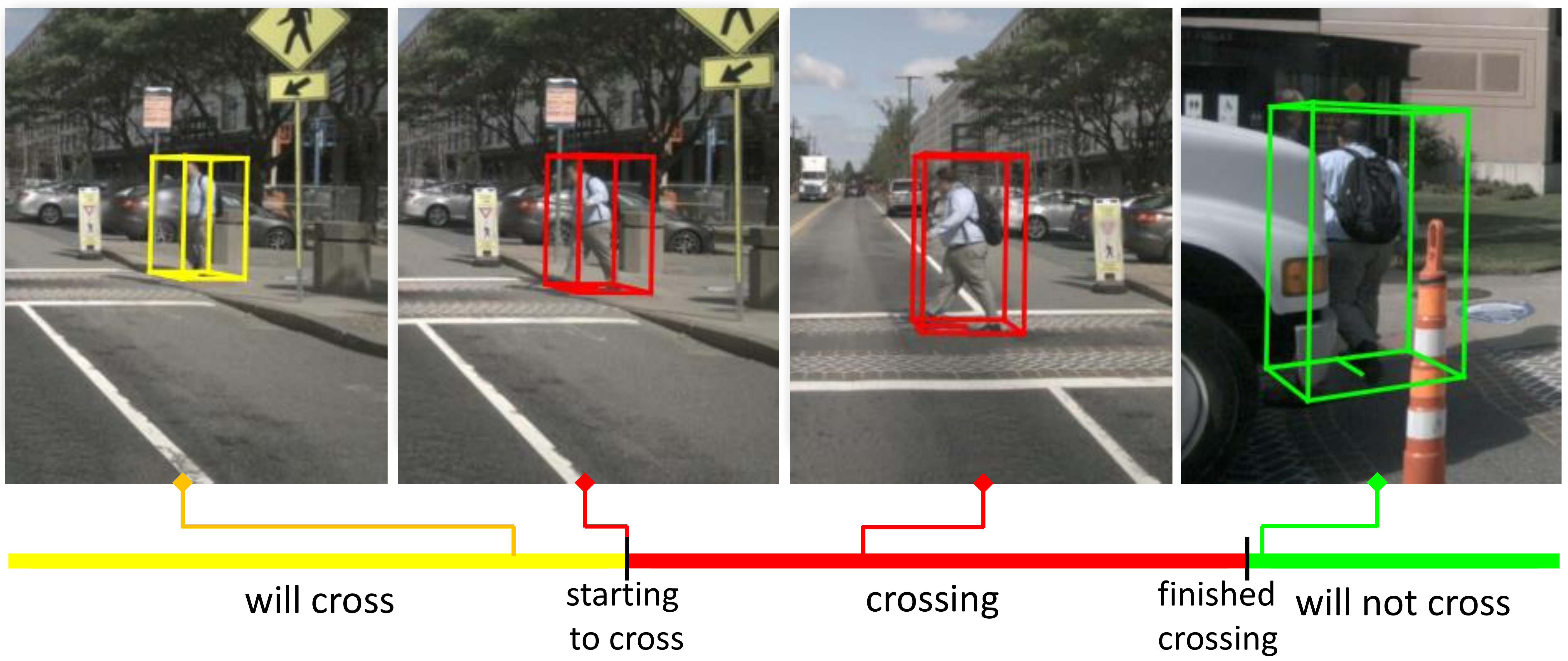}
    % \caption{An example of a pedestrian in PePScenes with bounding box and behavioural annotations. Green indicates a pedestrian that will not cross, yellow indicates a pedestrian before crossing, and red represents currently crossing.}
    \caption{An example of a pedestrian in PePScenes with bounding box and behavioural annotations. }
    \vspace{-1.5em}
    \label{fig:data_sample}
\end{figure}

The overall statistics of the proposed dataset can be found in Table \ref{data_stats}. Figure \ref{fig:data_sample} shows an example of a pedestrian with spatial and behavioural annotations.

\begin{table}[htp]
    \centering
    % \vspace{-0.25cm}
    \caption{The overall statistics of the annotations. The numbers under \textit{New} refer to the newly added annotations and under \textit{Original} the existing nuScenes annotations.}
    \begin{tabular}{l|cc|c}
    \multicolumn{1}{c|}{Annt.} & \textit{New} & \multicolumn{1}{l|}{\textit{Original}} & Total \\ \hline \hline
    \# Ped. with Beh.       & 719   & -     & 719   \\
    \# Cross. Peds.         & 149   & -     & 149   \\
    \# Non-cross Peds.      & 570   & -     & 570   \\
    \# Per-frame beh. annt. & 63.4k & -     & 63.4k \\
    \# Ped. box annt.       & 845k  & 222k  & 1.06M \\
    \# Other box annt.      & 3.58M & 944k  & 4.52M \\
    Annt. frame rate        & \multicolumn{1}{l}{10Hz} & 2Hz & 10Hz
    \end{tabular}
    \label{data_stats}
    \vspace{-0.5cm}
\end{table}

\section{Evaluation\vspace{-0.2\baselineskip}}

\subsection{Implementation\vspace{-0.2\baselineskip}}

\textbf{Training}.
We trained our model, Graph-SIM (Graph-based Spatiotemporal Interaction Modelling), using the Adam optimizer \cite{Kingma_2015_ICLR}, learning rate of 0.000002, and batch size of 16, for 10 epochs. To account for class imbalance, we weighted the classes based on the ratio of positive to negative samples. 

\textbf{Data}.
We divided the pedestrians into training and test sets, using 70\% and 30\% of the data respectively, while ensuring a consistent ratio of positive to negative pedestrians. 

As done in \cite{Rasouli_2019_arXiv}, for positive pedestrians, we keep only the observations up to the frame where crossing begins. For negative pedestrians, we discard the segment of observations that occur after the pedestrian leaves the view of the front cameras (left, center, and right). 

We choose an observation length of 5 frames (0.5 seconds), sampled from one to two seconds before the crossing event, with an overlap of 50\% between sequences.

\textbf{Metrics}.
For evaluation, we use common binary classification metrics as in \cite{Rasouli_2019_arXiv}, including \textit{accuracy}, Area Under the Curve (\textit{AUC}), \textit{F1}, and \textit{precision}. Given that the number of positive and negative samples are imbalanced in PePScenes, it is possible for the model to achieve high accuracy while favouring one class. Thus, we report all these metrics to provide a holistic view of model performance. 

\subsection{Models\vspace{-0.2\baselineskip}}
\label{sect:models}
We compare the performance of our proposed method to two state-of-the-art models.

\textbf{Stacked with Fusion GRU (SF-GRU) \cite{Rasouli_2019_arXiv}}.
This pedestrian action prediction model has a multi-level recurrent architecture that encodes and infuses five modalities of data gradually -- pedestrians' appearance, surrounding context, and poses, 2D bounding box coordinates, and the ego-vehicle speed. We modify this to use 3D global coordinates instead of 2D bounding boxes.

\textbf{Social-STGCNN \cite{Mohamed_2020_CVPR}}.
This is a graph-based pedestrian trajectory prediction model, which we modify for action prediction using 
our architecture, as in Fig. \ref{fig:architecture}. Just like \cite{Mohamed_2020_CVPR}, 1) we use only bird's-eye-view locations of road users for graph node representations and 2) edge weights are computed as $1/d$, where $d$ is the distance between each object and the target pedestrian. With this, we use the pedestrian-centric star graph component of our model (without clustering).

\subsection{Prediction Results on PePScenes\vspace{-0.2\baselineskip}}
The models described in the previous section were evaluated along with three versions of Graph-SIM -- graph modelling with 1) the target pedestrian's dynamics, 2) the ego-vehicle's dynamics, and 3) both the target pedestrian and ego-vehicle dynamics. 

As shown in Table \ref{tab:models}, the complete model consisting of both ego-vehicle and pedestrian dynamics outperforms Social-STGCNN in all metrics, particularly F1 and precision, by 20\% and 29\% respectively. It is also apparent from the results that ego-vehicle motion has a more dominant impact on the performance in comparison to pedestrian dynamics, which are already encoded in the graph. 

\begin{table}[htp]
    \centering
    \vspace{-0.25cm}
    \caption{Evaluation results on PePScenes.}
    \begin{tabular}{ll|ccccc}
        \multicolumn{2}{l|}{Method} & Acc & AUC & F1 & Prec \\ \hline  \hline
        \multicolumn{2}{l|}{SF-GRU \cite{Rasouli_2019_arXiv}} & 0.867 & 0.683 & 0.506 & 0.672 \\ 
        \multicolumn{2}{l|}{Social-STGCNN \cite{Mohamed_2020_CVPR}} & 0.874 & 0.762 & 0.613 & 0.633 \\ \hline
        \multirow{3}{*}{Graph-SIM} 
            & Graph+Ped. & 0.805 & 0.608 & 0.349 & 0.395 \\
            & Graph+Veh. & 0.916 & 0.821 & 0.730 & 0.793 \\
			& \textbf{Complete}  &  \textbf{0.944} & \textbf{0.858} & \textbf{0.814} & \textbf{0.921} \\
    \end{tabular}
    \label{tab:models}
    \vspace{-0.7cm}
\end{table}

\subsection{Ablation\vspace{-0.2\baselineskip}}

\textbf{Graph-based Interaction Modelling.} 
In this section, we evaluate different variations of our proposed star graph (S-Graph) approach and compare the results with Social-STGCNN. As illustrated in Table \ref{tab:graph_ablation}, S-Graph using relative importance modelling outperforms Social-STGCNN. 

In addition, clustering further improves the results, as it captures the social forces and overall behaviour of road users in groups, as highlighted in past studies \cite{Rasouli_2019_ITS}. We observe that including pedestrian orientation in clustering captures their group membership more effectively than with proximity alone. For example, two people might be close to each other while moving towards opposite directions. This resulted in the best overall performance, particularly improving precision by 5\%.

\addtolength{\tabcolsep}{-2pt}
\begin{table}[htp]
    \centering
    \vspace{-0.25cm}
    \caption{Ablation study on different graph modelling techniques. RI stands for \textit{Relative Importance}.}
    \begin{tabular}{l|ccccc}
        \multicolumn{1}{l|}{Graph Component}  & Acc & AUC & F1 & Prec \\ \hline \hline
        \multicolumn{1}{l|}{Social-STGCNN \cite{Mohamed_2020_CVPR}} & 0.874 & 0.762 & 0.613 & 0.633 \\
        \multicolumn{1}{l|}{S-Graph+RI} & 0.873 & 0.861 & 0.689 & 0.583 \\
        \multicolumn{1}{l|}{S-Graph+RI+Clust. w/o Ped. Orient. } & 0.941 & \textbf{0.869} & 0.811 & 0.869 \\ 
        \multicolumn{1}{l|}{\textbf{S-Graph+RI+Clust.}} & \textbf{0.944} & 0.858 & \textbf{0.814} & \textbf{0.921} \\
    \end{tabular}
    \label{tab:graph_ablation}
    \vspace{-0.3cm}
\end{table}
\addtolength{\tabcolsep}{+2pt}

\textbf{Target Pedestrian and Ego-Vehicle Dynamics.} 
We examine the contribution of different modalities in our dynamics encoding. The results are shown in Table \ref{tab:dynamic_ablation}. 

As expected, with the combination of pedestrian location and velocity, the model performs better. In isolation, these features are insufficient to achieve a high accuracy because location and direction of motion are equally important to capture pedestrian dynamics. By including ego-vehicle dynamics, performance improves. This is primarily because ego-vehicle motion can directly impact pedestrians behaviour.

\begin{table}[htp]
    \centering
    \vspace{-0.2cm}
    \caption{Ablation study for the target pedestrian and ego-vehicle dynamics.}
    \begin{tabular}{l|ccccc}
        \multicolumn{1}{l|}{Input Features}  & Acc & AUC & F1 & Prec \\ \hline \hline
        \multicolumn{1}{l|}{Ped. loc.} & 0.763 & 0.500 & 0.128 & 0.167 \\
        \multicolumn{1}{l|}{Ped. vel.} & 0.565 & \textbf{0.614} & 0.346 & 0.232 \\
        \multicolumn{1}{l|}{Ped. loc./vel.} & \textbf{0.805} & 0.608 & \textbf{0.349} & \textbf{0.395} \\ \hline
        \multicolumn{1}{l|}{Ped. loc./vel. + Veh. loc.} & 0.735 & 0.587 & 0.315 & 0.278 \\
        \multicolumn{1}{l|}{Ped. loc./vel. + Veh. vel.} & 0.820 & 0.609 & 0.352 & 0.444 \\
        \multicolumn{1}{l|}{\textbf{Ped. loc./vel. + Veh. loc./vel.}} & \textbf{0.944} & \textbf{0.858} & \textbf{0.814} & \textbf{0.921} \\
    \end{tabular}
    \label{tab:dynamic_ablation}
    \vspace{-0.5cm}
\end{table}

\section{Conclusion\vspace{-0.2\baselineskip}}
In this work, we proposed a novel model for pedestrian crossing prediction. Our method uses a graph-based structure to model interactions between target pedestrians and their surroundings. The nodes in the graph are represented by road users, clustered according to their global bird's-eye-view coordinates and orientations. The edges are weighted by calculating the relative importance of the road users to the target pedestrian. 

In addition, we proposed a new dataset by providing additional 3D bounding box and behavioural annotations to the existing nuScenes dataset.
Using this data, we showed that our model exceeds the performance of existing approaches by over 15\% on average across various metrics. 

\bibliographystyle{IEEEtran}
\bibliography{refs}

\begin{thebibliography}{10}
\providecommand{\url}[1]{#1}
\csname url@rmstyle\endcsname
\providecommand{\newblock}{\relax}
\providecommand{\bibinfo}[2]{#2}
\providecommand\BIBentrySTDinterwordspacing{\spaceskip=0pt\relax}
\providecommand\BIBentryALTinterwordstretchfactor{4}
\providecommand\BIBentryALTinterwordspacing{\spaceskip=\fontdimen2\font plus
\BIBentryALTinterwordstretchfactor\fontdimen3\font minus
  \fontdimen4\font\relax}
\providecommand\BIBforeignlanguage[2]{{%
\expandafter\ifx\csname l@#1\endcsname\relax
\typeout{** WARNING: IEEEtran.bst: No hyphenation pattern has been}%
\typeout{** loaded for the language `#1'. Using the pattern for}%
\typeout{** the default language instead.}%
\else
\language=\csname l@#1\endcsname
\fi
#2}}

\bibitem{Alahi_2016_CVPR}
A.~Alahi, K.~Goel, V.~Ramanathan, A.~Robicquet, L.~Fei-Fei, and S.~Savarese,
  ``Social lstm: Human trajectory prediction in crowded spaces,'' in
  \emph{CVPR}, 2016.

\bibitem{Bhattacharyya_2018_CVPR}
A.~Bhattacharyya, M.~Fritz, and B.~Schiele, ``Long-term on-board prediction of
  people in traffic scenes under uncertainty,'' in \emph{CVPR}, 2018.

\bibitem{Kooji_2014_ECCV}
J.~F.~P. Kooij, N.~Schneider, F.~Flohr, and D.~Gavrila, ``Context-based
  pedestrian path prediction,'' in \emph{ECCV}, 2014.

\bibitem{Zhao_2020_ArXiv}
H.~Zhao, J.~Gao, T.~Lan, C.~Sun, B.~Sapp, B.~Varadarajan, Y.~Shen, Y.~Chai,
  C.~Schmid, C.~Li, and D.~Anguelov, ``Tnt: Target-driven trajectory
  prediction,'' in \emph{CoRL}, 2020.

\bibitem{Yu_2020_ECCV}
C.~Yu, X.~Ma, J.~Ren, H.~Zhao, and S.~Yi, ``Spatio-temporal graph transformer
  networks for pedestrian trajectory prediction,'' in \emph{ECCV}, 2020.

\bibitem{Haddad_2020_arXiv}
S.~Haddad, M.~Wu, H.~Wei, and S.~Lam, ``Situation-aware pedestrian trajectory
  prediction with spatio-temporal attention model,'' in \emph{WACV}, 2019.

\bibitem{Manh_2019_ISVC}
M.~Huynh and G.~Alaghband, ``{Trajectory Prediction by Coupling Scene-LSTM with
  Human Movement LSTM},'' in \emph{International Symposium on Visual Computing
  (ISVC)}, 2019.

\bibitem{Fang_2018_IV}
Z.~Fang and A.~M. Lo\'pez, ``Is the pedestrian going to cross? answering by 2d
  pose estimation,'' in \emph{IEEE IV}, 2018.

\bibitem{Stip}
B.~Liu, E.~Adeli, Z.~Cao, K.-H. Lee, A.~Shenoi, A.~Gaidon, and J.~C. Niebles,
  ``Spatiotemporal relationship reasoning for pedestrian intent prediction,''
  \emph{IEEE RA-L}, 2020.

\bibitem{Jaad}
A.~Rasouli, I.~Kotseruba, and J.~K. Tsotsos, ``Are they going to cross? a
  benchmark dataset and baseline for pedestrian crosswalk behavior,'' in
  \emph{ICCVW}, 2017.

\bibitem{Sun_2020_CVPR}
J.~{Sun}, Q.~{Jiang}, and C.~{Lu}, ``Recursive social behavior graph for
  trajectory prediction,'' in \emph{CVPR}, 2020.

\bibitem{Mohamed_2020_CVPR}
A.~Mohamed, K.~Qian, M.~Elhoseiny, and C.~Claudel, ``Social-stgcnn: A social
  spatio-temporal graph convolutional neural network for human trajectory
  prediction,'' in \emph{CVPR}, 2020.

\bibitem{Pie}
A.~Rasouli, I.~Kotseruba, T.~Kunic, and J.~Tsotsos, ``Pie: A large-scale
  dataset and models for pedestrian intention estimation and trajectory
  prediction,'' in \emph{ICCV}, 2019.

\bibitem{Titan}
S.~Malla, B.~Dariush, and C.~Choi, ``Titan: Future forecast using action
  priors,'' in \emph{CVPR}, 2020.

\bibitem{Viena2}
M.~S. Aliakbarian, F.~Saleh, M.~Salzmann, B.~Fernando, L.~Petersson, and
  L.~Andersson, ``Viena2: A driving anticipation dataset,'' in \emph{ACCV},
  2018.

\bibitem{Caesar_2020_CVPR}
H.~Caesar, V.~Bankiti, A.~H. Lang, S.~Vora, V.~E. Liong, Q.~Xu, A.~Krishnan,
  Y.~Pan, G.~Baldan, and O.~Beijbom, ``nuscenes: A multimodal dataset for
  autonomous driving,'' in \emph{CVPR}, 2020.

\bibitem{Sun_2020_CVPR_2}
H.~Sun, Z.~Zhao, and Z.~He, ``Reciprocal learning networks for human trajectory
  prediction,'' in \emph{CVPR}, 2020.

\bibitem{Kosaraju_2019_NIPS}
V.~Kosaraju, A.~Sadeghian, R.~Mart\'{\i}n-Mart\'{\i}n, I.~Reid, H.~Rezatofighi,
  and S.~Savarese, ``Social-bigat: Multimodal trajectory forecasting using
  bicycle-gan and graph attention networks,'' in \emph{NeurIPS}, 2019.

\bibitem{Zhang_2020_CVPR}
Z.~Zhang, J.~Gao, J.~Mao, Y.~Liu, D.~Anguelov, and C.~Li, ``Stinet:
  Spatio-temporal-interactive network for pedestrian detection and trajectory
  prediction,'' in \emph{CVPR}, 2020.

\bibitem{Park_2020_ECCV}
S.~H. Park, G.~Lee, M.~Bhat, J.~Seo, M.~Kang, J.~Francis, A.~R. Jadhav, P.~P.
  Liang, and L.-P. Morency, ``Diverse and admissible trajectory forecasting
  through multimodal context understanding,'' in \emph{ECCV}, 2020.

\bibitem{Chaabane_2020_WACV}
M.~Chaabane, A.~Trabelsi, N.~Blanchard, and J.~Beveridge, ``Looking ahead:
  Anticipating pedestrians crossing with future frames prediction,'' in
  \emph{WACV}, 2020.

\bibitem{Liang_2019_CVPR}
J.~Liang, L.~Jiang, J.~C. Niebles, A.~G. Hauptmann, and L.~Fei-Fei, ``Peeking
  into the future: Predicting future person activities and locations in
  videos,'' in \emph{CVPR}, 2019.

\bibitem{Gujjar_2019_ICRA}
P.~Gujjar and R.~Vaughan, ``Classifying pedestrian actions in advance using
  predicted video of urban driving scenes,'' in \emph{ICRA}, 2019.

\bibitem{Saleh_2019_ICRA}
K.~Saleh, M.~Hossny, and S.~Nahavandi, ``Real-time intent prediction of
  pedestrians for autonomous ground vehicles via spatio-temporal densenet,'' in
  \emph{ICRA}, 2019.

\bibitem{Rasouli_2019_arXiv}
A.~Rasouli, I.~Kotseruba, and J.~K. Tsotsos, ``Pedestrian action anticipation
  using contextual feature fusion in stacked rnns,'' in \emph{BMVC}, 2019.

\bibitem{Gupta_2018_CVE}
A.~Gupta, J.~Johnson, L.~Fei-Fei, S.~Savarese, and A.~Alahi, ``Social gan:
  Socially acceptable trajectories with generative adversarial networks,'' in
  \emph{CVPR}, 2018.

\bibitem{Mangalam_2020_ECCV}
K.~Mangalam, H.~Girase, S.~Agarwal, K.-H. Lee, E.~Adeli, J.~Malik, and
  A.~Gaidon, ``It is not the journey but the destination: Endpoint conditioned
  trajectory prediction,'' in \emph{ECCV}, 2020.

\bibitem{Vemula_2018_ICRA}
A.~Vemula, K.~Muelling, and J.~Oh, ``Social attention: Modeling attention in
  human crowds,'' in \emph{ICRA}, 2018.

\bibitem{Sadeghian_2019_CVPR}
A.~Sadeghian, V.~Kosaraju, A.~Sadeghian, N.~Hirose, H.~Rezatofighi, and
  S.~Savarese, ``Sophie: An attentive gan for predicting paths compliant to
  social and physical constraints,'' in \emph{CVPR}, 2019.

\bibitem{Rasouli_2019_ITS}
A.~{Rasouli} and J.~K. {Tsotsos}, ``Autonomous vehicles that interact with
  pedestrians: A survey of theory and practice,'' \emph{IEEE T-ITS}, 2020.

\bibitem{Ester_1996_KDD}
M.~Ester, H.-P. Kriegel, J.~Sander, and X.~Xu, ``A density-based algorithm for
  discovering clusters in large spatial databases with noise,'' in
  \emph{International Conference on Knowledge Discovery and Data Mining (KDD)},
  1996.

\bibitem{Lloyd_1982_IT}
S.~P. Lloyd, ``Least squares quantization in pcm.'' \emph{IEEE T-IT}, 1982.

\bibitem{kipf}
T.~Kipf and M.~Welling, ``Semi-supervised classification with graph
  convolutional networks,'' in \emph{ICLR}, 2017.

\bibitem{Prelu}
K.~He, X.~Zhang, S.~Ren, and J.~Sun, ``Delving deep into rectifiers: Surpassing
  human-level performance on imagenet classification,'' in \emph{ICCV}, 2015.

\bibitem{Lin_2017_CVPR}
T.-Y. Lin, P.~Goyal, R.~Girshick, K.~He, and P.~Doll{\'a}r, ``Focal loss for
  dense object detection,'' in \emph{CVPR}, 2017.

\bibitem{Lin_2014_ECCV}
T.-Y. Lin, M.~Maire, S.~Belongie, J.~Hays, P.~Perona, D.~Ramanan,
  P.~Doll{\'a}r, and C.~L. Zitnick, ``Microsoft {COCO}: Common objects in
  context,'' in \emph{ECCV}, 2014.

\bibitem{Kingma_2015_ICLR}
D.~P. Kingma and J.~Ba, ``Adam: {A} method for stochastic optimization,'' in
  \emph{ICLR}, 2015.

\end{thebibliography}

\end{document}